# Multi-document Biography Summarization

Liang Zhou, Miruna Ticrea, Eduard Hovy
University of Southern California
Information Sciences Institute
4676 Admiralty Way
Marina del Rey, CA 90292-6695
{liangz, miruna, hovy} @isi.edu

## Abstract

In this paper we describe a biography summarization system using sentence classification and ideas from information retrieval. Although the individual techniques are not new, assembling and applying them to generate multi-document biographies is new. Our system was evaluated in DUC2004. It is among the top performers in task 5–short summaries focused by person questions.

## 1    Introduction

Automatic text summarization is one form of information management. It is described as selecting a subset of sentences from a document that is in size a small percentage of the original and yet is just as informative. Summaries can serve as surrogates of the full texts in the context of Information Retrieval (IR). Summaries are created from two types of text sources, a single document or a set of documents. Multi-document summarization (MDS) is a natural and more elaborative extension of single-document summarization, and poses additional difficulties on algorithm design. Various kinds of summaries fall into two broad categories: generic summaries are the direct derivatives of the source texts; special-interest summaries are generated in response to queries or topic-oriented questions.

One important application of special-interest MDS systems is creating biographies to answer questions like "who is Kofi Annan?". This task would be tedious for humans to perform in situations where the information related to the person is deeply and sparsely buried in large quantity of news texts that are not obviously related. This paper describes a MDS biography system that responds to the "who is" questions by identifying information about the person-in-question using IR and classification techniques, and creates multi-document biographical summaries. The overall system design is shown in Figure 1.

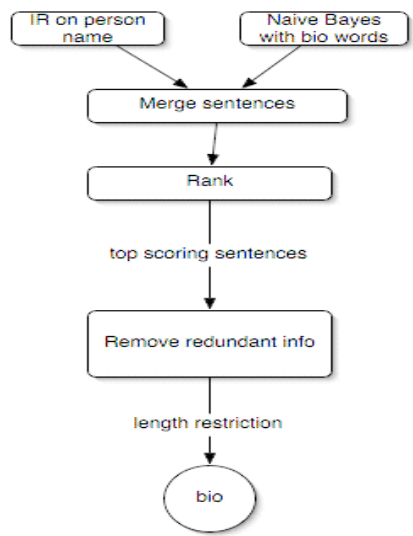

Figure 1. Overall design of the biography summarization system.

To determine what and how sentences are selected and ranked, a simple IR method and experimental classification methods both contributed. The set of top-scoring sentences, after redundancy removal, is the resulting biography. As yet, the system contains no inter-sentence 'smoothing' stage.

In this paper, work in related areas is discussed in Section 2; a description of our biography corpus used for training and testing the classification component is in Section 3; Section 4 explains the need and the process of classifying sentences according to their biographical state; the application of the classification method in biography extraction/summarization is described in Section 5; an accompanying evaluation on the quality of the biography summaries is shown in Section 6; and future work is outlined in Section 7.

## 2    Recent Developments

Two trends have dominated automatic summarization research (Mani, 2001). One is the work focusing on generating summaries by extraction, which is finding a subset of the document that is indicative of its contents (Kupiec et al., 1995) using "shallow" linguistic analysis and statistics. The other influence is the exploration of

"deeper" knowledge-based methods for condensing information. Knight and Marcu (2000) equate summarization with compression at sentence level to achieve grammaticality and information capture, and push a step beyond sentence extraction. Many systems use machine-learning methods to learn from readily aligned corpora of scientific articles and their corresponding abstracts. Zhou and Hovy (2003) show a summarization system trained from automatically obtained text-summary alignments obeying the chronological occurrences of news events.

MDS poses more challenges in assessing similarities and differences among the set of documents. The simple idea of extract-and-concatenate does not respond to problems arisen from coherence and cohesion. Barzilay et al. (1999) introduce a combination of extracted similar phrases and a reformulation through sentence generation. Lin and Hovy (2002) apply a collection of known single-document summarization techniques, cooperating positional and topical information, clustering, etc., and extend them to perform MDS.

While many have suggested that conventional MDS systems can be applied to biography generation directly, Mani (2001) illustrates that the added functionality of biographical MDS comes at the expense of a substantial increase in system complexity and is somewhat beyond the capabilities of present day MDS systems. The discussion was based in part on the only known MDS biography system (Schiffman et al., 2001) that uses corpus statistics along with linguistic knowledge to select and merge description of people in news. The focus of this work was on synthesizing succinct descriptions of people by merging appositives from semantic processing using WordNet (Miller, 1995).

## 3 Corpus Description

In order to extract information that is related to a person from a large set of news texts written not exclusively about this person, we need to identify attributes shared among biographies.

Biographies share certain standard components. We annotated a corpus of 130 biographies of 12 people (activists, artists, leaders, politicians, scientists, terrorists, etc.). We found 9 common elements: *bio* (info on birth and death), *fame factor, personality, personal, social, education, nationality, scandal,* and *work*. Collected biographies are appropriately marked at clause-level with one of the nine tags in XML format, for example:

```
Martin Luther King <nationality>
was born in Atlanta, Georgia
</nationality>. … He <bio>was
assassinated on April 4, 1968
</bio>. … King <education> entered
the Boston University as a
doctoral student </education>. …
```

In all, 3579 biography-related phrases were identified and recorded for the collection, among them 321 *bio*, 423 *fame*, 114 *personality*, 465 *personal*, 293 *social*, 246 *education*, 95 *nationality*, 292 *scandal*, and 1330 *work*. We then used 100 biographies for training and 30 for testing the classification module.

## 4 Sentence Classification

Relating to human practice on summarizing, three main points are relevant to aid the automation process (Spärck Jones, 1993). The first is a strong emphasis on particular purposes, e.g., abstracting or extracting articles of particular genres. The second is the drafting, writing, and revision cycle in constructing a summary. Essentially as a consequence of these first two points, the summarizing process can be guided by the use of checklists. The idea of a checklist is especially useful for the purpose of generating biographical summaries because a complete biography should contain various aspects of a person's life. From a careful analysis conducted while constructing the biography corpus, we believe that the checklist is shared and common among all persons in question, and consists the 9 biographical elements introduced in Section 3.

The task of fulfilling the biography checklist becomes a classification problem. Classification is defined as a task of classifying examples into one of a discrete set of possible categories (Mitchell, 1997). Text categorization techniques have been used extensively to improve the efficiency on information retrieval and organization. Here the problem is that sentences, from a set of documents, need to be categorized into different biography-related classes.

### 4.1 Task Definitions

We designed two classification tasks:

1) 10 -Class: Given one or more texts about a person, the module must categorize each sentence into one of ten classes. The classes are the 9 biographical elements plus a class called *none* that collects all sentences without biographical information. This fine-grained classification task will be beneficial in generating comprehensive biographies on people of interest. The classes are:

```
bio
fame
personality
social
education
nationality
scandal
personal
work
none
```

2) 2-Class: The module must make a binary decision of whether the sentence should be included in a biography summary. The classes are:

```
bio
none
```

The label *bio* appears in both task definitions but bears different meanings. Under 10-Class, class *bio* contains information on a person's birth or death, and under 2-Class it sums up all 9 biographical elements from the 10-Class.

## 4.2   Machine Learning Methods

We experimented with three machine learning methods for classifying sentences.

**Naïve Bayes**

The Naïve Bayes classifier is among the most effective algorithms known for learning to classify text documents (Mitchell, 1997), calculating explicit probabilities for hypotheses. Using $k$ features $F_j: j = 1, ..., k$, we assign to a given sentence $S$ the class $C$:

$$C = \arg\max_C P(C \mid F_1, F_2, ..., F_k)$$

It can be expressed using Bayes' rule, as (Kupiec et al., 1995):

$$P(S \in C \mid F_1, F_2, ...F_k) = \frac{P(F_1, F_2, ...F_j \mid S \in C) \bullet P(S \in C)}{P(F_1, F_2, ...F_k)}$$

Assuming statistical independence of the features:

$$P(S \in C \mid F_1, F_2, ...F_k) = \frac{\prod_{j=1}^{k} P(F_j \mid S \in C) \bullet P(S \in C)}{\prod_{j=1}^{k} P(F_j)}$$

Since $P(F_j)$ has no role in selecting $C$:

$$P(S \in C \mid F_1, F_2, ...F_k) = \prod_{j=1}^{k} P(F_j \mid S \in C) \bullet P(S \in C)$$

We trained on the relative frequency of $P(F_j \mid S \in C)$ and $P(S \in C)$, with add-one smoothing. This method was used in classifying both the 10-Class and the 2-Class tasks.

**Support Vector Machine**

Support Vector Machines (SVMs) have been shown to be an effective classifier in text categorization. We extend the idea of classifying documents to predefined categories to classifying sentences into one of the two biography categories defined by the 2-Class task. Sentences are categorized based on their biographical saliency (a percentage of clearly identified biography words) and their non-biographical saliency (a percentage of clearly identified non-biography words). We used LIBSVM (Chang and Lin, 2003) for training and testing.

**Decision Tree (4.5)**

In addition to SVM, we also used a decision-tree algorithm, C4.5 (Quinlan, 1993), with the same training and testing data as SVM.

## 4.3   Classification Results

The lower performance bound is set by a baseline system that randomly assigns a biographical class given a sentence, for both 10-Class and 2-Class. 2599 testing sentences are from 30 unseen documents.

**10-Class Classification**

The Naïve Bayes classifier was used to perform the 10-Class task. Table 1 shows its performance with various features.

| Features | Precision/Recall (%) |
|---|---|
| Baseline | 10.04 |
| Unigram | 69.41 |
| Unigram + POS | 70.64 |
| Stem + POS | 68.10 |
| Expanded unigram (bio words + wordnet hypernyms) | 60.98 |
| Bigram | 69.45 |
| Relaxed unigram | 70.41 |
| Relaxed unigram + POS | 71.53 |
| Manually identified "work" words added | 68.37 |

Table 1. Performance of 10-Class sentence classification, using Naïve Bayes Classifier.

Part-of-speech (POS) information (Brill, 1995) and word stems (Lovins, 1968) were used in some feature sets.

We bootstrapped 10395 more biography-indicating words by recording the immediate hypernyms, using WordNet (Fellbaum, 1998), of the words collected from the controlled biography corpus described in Section 3. These words are called *Expanded Unigrams* and their frequency scores are reduced to a fraction of the original word's frequency score.

Some sentences in the testing set were labeled with multiple biography classes due to the fact that the original corpus was annotated at clause level. Since the classification was done at sentence level, we relaxed the matching/evaluating program allowing a hit when any of the several classes was matched. This is shown in Table 1 as the *Relaxed* cases.

A closer look at the instances where the false negatives occur indicates that the classifier mislabeled instances of class *work* as instances of class *none*. To correct this error, we created a list of 5516 *work* specific words hoping that this would set a clearer boundary between the two classes. However performance did not improve.

**2-Class Classification**

All three machine learning methods were evaluated in classifying among 2 classes. The results are shown in Table 2. The testing data is slightly skewed with 68% of the sentences being *none*.

| Features | | Precision/Recall (%) |
|---|---|---|
| Baseline | | 49.37 |
| Naïve Bayes | Unigram (from bio phrases) | 82.42 |
| | Bigram | 47.98 |
| | Unigram (from bio sentences) | 68.23 |
| | Unigram + POS (bio sentences) | 65.06 |
| SVM | | 74.47 |
| C4.5 | | 75.76 |

Table 2. Classification results on 2-Class using Naïve Bayes, SVM, and C4.5.

In addition to using marked biographical phrases as training data, we also expanded the marking/tagging perimeter to sentence boundaries. As shown in the table, this creates noise.

**5    Biography Extraction**

Biographical sentence classification module is only one of two components that supply the overall system with usable biographical contents, and is followed by other stages of processing (see system design in Figure 1). We discuss the other modules next.

**5.1    Name-filter**

A filter scans through all documents in the set, eliminating sentences that are direct quotes, dialogues, and too short (under 5 words). Person-oriented sentences containing any variation (first name only, last name only, and the full name) of the person's name are kept for subsequent steps.

Sentences classified as biography-worthy are merged with the name-filtered sentences with duplicates eliminated.

**5.2    Sentence Ranking**

An essential capability of a multi-document summarizer is to combine text passages in a useful manner for the reader (Goldstein et al., 2000). This includes a sentence ordering parameter (Mani, 2001). Each of the sentences selected by the name-filter and the biography classifier is either related to the person-in-question via some news event or referred to as part of this person's biographical profile, or both. We need a mechanism that will select sentences that are of informative significance within the source document set. Using inverse-term-frequency (ITF), i.e. an estimation of information value, words with high information value (low ITF) are distinguished from those with low value (high ITF). A sorted list of words along with their ITF scores from a document set—topic ITFs—displays the important events, persons, etc., from this particular set of texts. This allows us to identify passages that are unusual with respect to the texts about the person.

However, we also need to identify passages that are unusual in general. We have to quantify how these important words compare to the rest of the world. The world is represented by 413307562 words from TREC-9 corpus (http://trec.nist.gov/data.html), with corresponding ITFs.

The overall informativeness of each word $w$ is:

$$C_w = \frac{d_{itf_w}}{W_{itf_w}}$$

where $d_{itf}$ is the document set ITF of word $w$ and $W_{itf}$ is the world ITF of $w$. A word that occurs frequently bears a lower $C_w$ score compared to a rarely used word (bearing high information value) with a higher $C_w$ score.

Top scoring sentences are then extracted according to:

$$C_s = \frac{\sum_{i=1}^{n} C_{w_i}}{len(s)}$$

The following is a set of sentences extracted according to the method described so far. The person-in-question is the famed cyclist Lance Armstrong.

```
1. Cycling helped him win his
battle with cancer, and
cancer helped him win the
Tour de France.
2. Armstrong underwent four
rounds of intense
chemotherapy.
3. The surgeries and
chemotherapy eliminated the
cancer, and Armstrong began
his cycling comeback.
4. The foundation supports
cancer patients and survivors
through education, awareness
and research.
5. He underwent months of
chemotherapy.
```

### 5.3 Redundancy Elimination

Summaries that emphasize the differences across documents while synthesizing common information would be the desirable final results. Removing similar information is part of all MDS systems. Redundancy is apparent in the Armstrong example from Section 5.2. To eliminate repetition while retaining interesting singletons, we modified (Marcu, 1999) so that an extract can be automatically generated by starting with a full text and systematically removing a sentence at a time as long as a stable semantic similarity with the original text is maintained. The original extraction algorithm was used to automatically create large volume of (*extract, abstract, text*) tuples for training extraction-based summarization systems with (*abstract, text*) input pairs.

Top-scoring sentences selected by the ranking mechanism described in Section 5.2 were the input to this component. The removal process was repeated until the desired summary length was achieved.

Applying this method to the Armstrong example, the result leaves only one sentence that contains the topics "chemotherapy" and "cancer". It chooses sentence 3, which is not bad, though sentence 1 might be preferable.

## 6 Evaluation

### 6.1 Overview

Extrinsic and intrinsic evaluations are the two classes of text summarization evaluation methods (Sparck Jones and Galliers, 1996). Measuring content coverage or summary informativeness is an approach commonly used for intrinsic evaluation. It measures how much source content was preserved in the summary.

A complete evaluation should include evaluations of the accuracy of components involved in the summarization process (Schiffman et al., 2001). Performance of the sentence classifier was shown in Section 4. Here we will show the performance of the resulting summaries.

### 6.2 Coverage Evaluation

An intrinsic evaluation of biography summary was recently conducted under the guidance of Document Understanding Conference (DUC2004) using the automatic summarization evaluation tool ROUGE (Recall-Oriented Understudy for Gisting Evaluation) by Lin and Hovy (2003). 50 TREC English document clusters, each containing on average 10 news articles, were the input to the system. Summary length was restricted to 665 bytes. Brute force truncation was applied on longer summaries.

The ROUGE-L metric is based on Longest Common Subsequence (LCS) overlap (Saggion et al., 2002). Figure 2 shows that our system (86) performs at an equivalent level with the best systems 9 and 10, that is, they both lie within our system's 95% upper confidence interval. The 2-class classification module was used in generating the answers. The figure also shows the performance data evaluated with lower and higher confidences set at 95%. The performance data are from official DUC results.

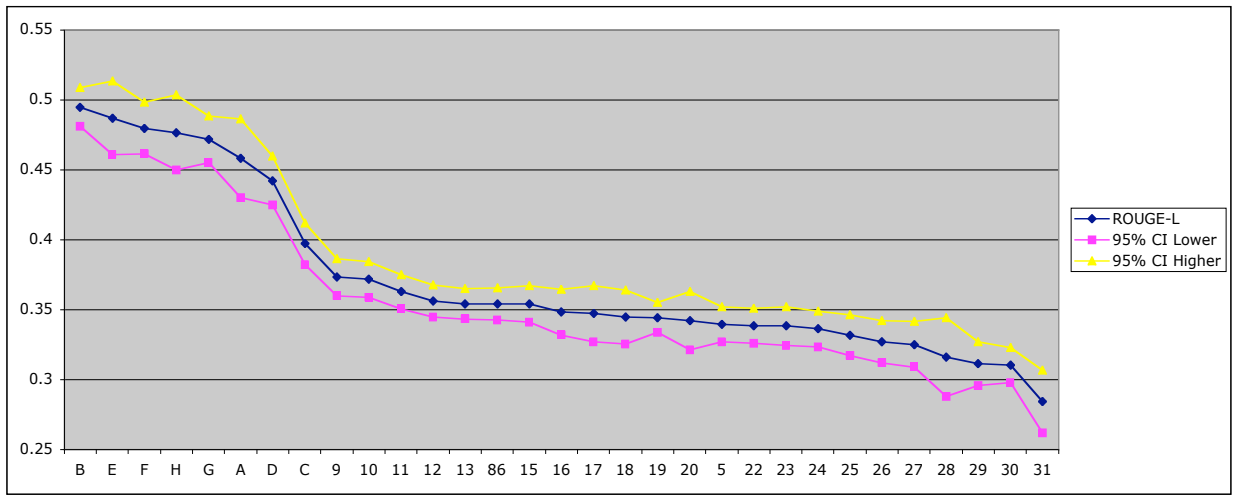

Figure 2. Official ROUGE performance results from DUC2004. Peer systems are labeled with numeric IDs. Humans are numbered A–H. 86 is our system with 2-class biography classification. Baseline is 5.

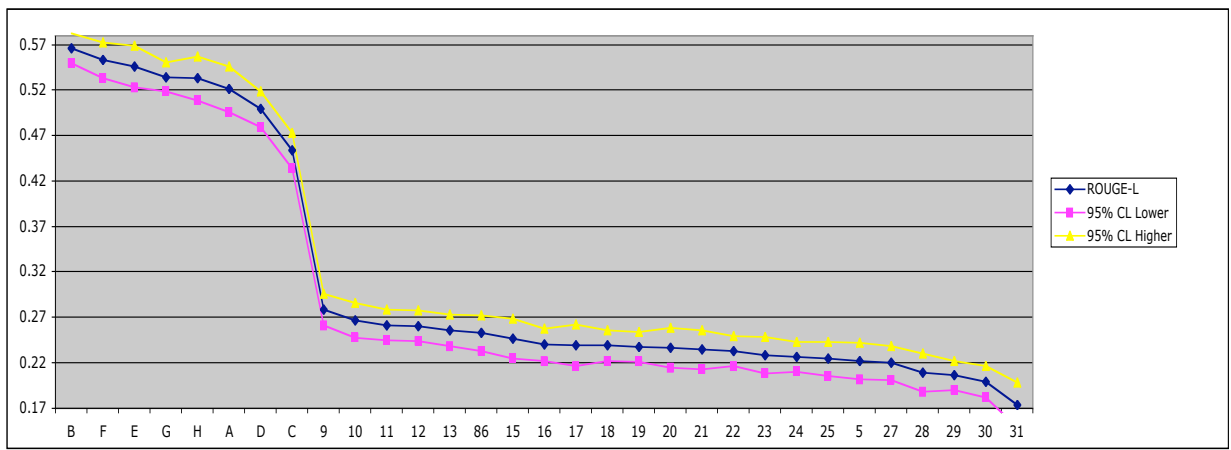

Figure 3. Unofficial ROUGE results. Humans are labeled A–H. Peer systems are labeled with numeric IDs. 86 is our system with 10-class biography classification. Baseline is 5.

Figure 3 shows the performance results of our system 86, using 10-class sentence classification, comparing to other systems from DUC by replicating the official evaluating process. Only system 9 performs slightly better with its score being higher than our system's 95% upper confidence interval.

A baseline system (5) that takes the first 665 bytes of the most recent text from the set as the resulting biography was also evaluated amongst the peer systems. Clearly, humans still perform at a level much superior to any system.

Measuring fluency and coherence is also important in reflecting the true quality of machine-generated summaries. There is no automated tool for this purpose currently. We plan to incorporate one for the future development of this work.

### 6.3 Discussion

N-gram recall scores are computed by ROUGE, in addition to ROUGE-L shown here. While cosine similarity and unigram and bigram overlap demonstrate a sufficient measure on content coverage, they are not sensitive on how information is sequenced in the text (Saggion et al., 2002). In evaluating and analyzing MDS results, metrics, such as ROUGE-L, that consider linguistic sequence are essential.

Radev and McKeown (1998) point out when summarizing interesting news events from multiple sources, one can expect reports with contradictory and redundant information. An intelligent summarizer should attain as much information as possible, combine it, and present it in the most concise form to the user. When we look at the different attributes in a person's life reported in news articles, a person is described by the job positions that he/she has held, by education institutions that he/she has attended, and etc. Those data are confirmed biographical information and do not bear the necessary contradiction associated with evolving news stories. However, we do feel the need to address and resolve discrepancies if we were to create comprehensive and detailed biographies on people-in-news since miscellaneous personal facts are often overlooked and told in conflicting reports. Misrepresented biographical information may well be controversies and may never be clarified. The *scandal* element from our corpus study (Section 3) is sufficient to identify information of the disputed kind.

Extraction-based MDS summarizers, such as this one, present the inherent problem of lacking the discourse-level fluency. While sentence ordering for single document summarization can be determined from the ordering of sentences in the input article, sentences extracted by a MDS system may be from different articles and thus need a strategy on ordering to produce a fluent surface summary (Barzilay et al., 2002). Previous summarization systems have used temporal sequence as the guideline on ordering. This is especially true in generating biographies where a person is represented by a sequence of events that occurred in his/her life. Barzilay et al. also introduced a combinational method with an alternative strategy that approximates the information relatedness across the input texts. We plan to use a fixed-form structure for the majority of answer construction, fitted for biographies only. This will be a top-down ordering strategy, contrary to the bottom-up algorithm shown by Barzilay et al.

## 7    Conclusion and Future Work

In this paper, we described a system that uses IR and text categorization techniques to provide summary-length answers to biographical questions. The core problem lies in extracting biography-related information from large volumes of news texts and composing them into fluent, concise, multi-document summaries. The summaries generated by the system address the question about the person, though not listing the chronological events occurring in this person's life due to the lack of background information in the news articles themselves. In order to obtain a "normal" biography, one should consult other means of information repositories.

Question: Who is Sir John Gielgud?

Answer: Sir John Gielgud, one of the great actors of the English stage who enthralled audiences for more than 70 years with his eloquent voice and consummate artistry, died Sunday at his home Gielgud's last major film role was as a surreal Prospero in Peter Greenaway's controversial Shakespearean rhapsody.

Above summary does not directly explain who the person-in-question is, but indirectly does so in explanatory sentences. We plan to investigate combining fixed-form and free-form structures in answer construction. The summary would include an introductory sentence of the form "x is <type/fame-category> ...", possibly through querying outside online resources. A main body would follow the introduction with an assembly of checklist items generated from the 10-Class classifier. A conclusion would contain open-ended items of special interest.

Furthermore, we would like to investigate compression strategies in creating summaries, specifically for biographies. Our biography corpus was tailored for this purpose and will be the starting point for further investigation.

**Acknowledgement**

We would like to thank Chin-Yew Lin from ISI for many insightful discussions on MDS, biography generation, and ROUGE.